\documentclass{article} 

\usepackage{iclr2027_conference,times}

\usepackage{amsmath,amsfonts,bm}

\def\eqref#1{equation~\ref{#1}}

\def\1{\bm{1}}

\DeclareMathAlphabet{\mathsfit}{\encodingdefault}{\sfdefault}{m}{sl}
\SetMathAlphabet{\mathsfit}{bold}{\encodingdefault}{\sfdefault}{bx}{n}

\usepackage{array}
\usepackage{adjustbox}
\usepackage{tabularx}
\usepackage{algorithm}
\usepackage{algpseudocode}

\usepackage{booktabs}
\usepackage{colortbl}
\usepackage{soul}
\usepackage{float}
\usepackage{graphicx}
\usepackage{hyperref}
\usepackage{placeins}
\usepackage{url}

\usepackage{wrapfig} 
\usepackage{booktabs} 
\usepackage{graphicx}

\newcolumntype{Y}{>{\raggedright\arraybackslash}X}
\newcommand{\StandardTable}{%
  \normalsize
  \setlength{\tabcolsep}{4pt}%
  \renewcommand{\arraystretch}{1.08}%
}
\newcommand{\CompactTable}{%
  \normalsize
  \setlength{\tabcolsep}{3.5pt}%
  \renewcommand{\arraystretch}{1.08}%
}

\title{Exploring a Layer-Wise Design Space for KV Cache Eviction}

\author{%
\textbf{Chao Fei}$^{1}$ \quad \textbf{Kaihua Liang}$^{1}$ \quad
\textbf{Hanzhi Hu}$^{1}$ \quad \textbf{Hongcheng Guo}$^{2}$\\
\textbf{Jian Weng}$^{1}$ \quad \textbf{Marco Canini}$^{1}$ \quad
\textbf{Panos Kalnis}$^{1}$\\
$^{1}$King Abdullah University of Science and Technology \quad
$^{2}$Fudan University\\
}

\iclrfinalcopy 
\begin{document}

\maketitle
\lhead{Preprint}

\begin{abstract}
KV cache eviction methods typically use a single retention-rule family throughout a model, making eviction-method identity a model-level design choice.
Yet Transformer layers differ substantially in their attention behavior, representations, and sensitivity to compression, suggesting that a uniform rule may overlook useful layer-wise structure.
This raises a basic question: should eviction methods themselves vary across layers?
We investigate this question by composing existing eviction methods across Transformer layers and systematically exploring the resulting layer-wise design space.
Using simple offline profiles, we construct fixed routes and study how their quality varies with method placement and cache budget.
On LongBench, heterogeneous routing improves performance on a majority of tasks over homogeneous policies at the same cache budget.
Even when method counts are held fixed, the profile-guided placement ranks second among 100 evaluated assignments, demonstrating that routing quality depends strongly on where methods are placed.
Moreover, the same fixed route outperforms the best of nine standalone baselines across all five tested cache budgets.
Together, these results establish layer-wise method composition as an exploitable, placement-sensitive design dimension for KV cache compression.
Code is available \href{https://anonymous.4open.science/r/PolyKV-iclr/README.md}{here}.
\end{abstract}

\section{Introduction}

Long-context inference powers increasingly capable LLM applications, including retrieval-augmented generation and language agents~\citep{rag,react}.
Yet its KV cache grows with context length, consuming substantial GPU memory and limiting the number of requests that can be served in parallel~\citep{h2o,pagedattention}.
KV cache eviction addresses this bottleneck by retaining only the historical states most likely to remain useful~\citep{h2o,scissorhands,streamingllm,snapkv}.
Existing methods encode distinct retention biases, but representative evaluations typically select one retention-rule family for the entire model.
Although layer-local scores can lead the same rule to retain different positions at different layers, the identity of the rule itself remains a model-level choice.

\begin{figure}[!t]
\centering
\includegraphics[width=\linewidth]{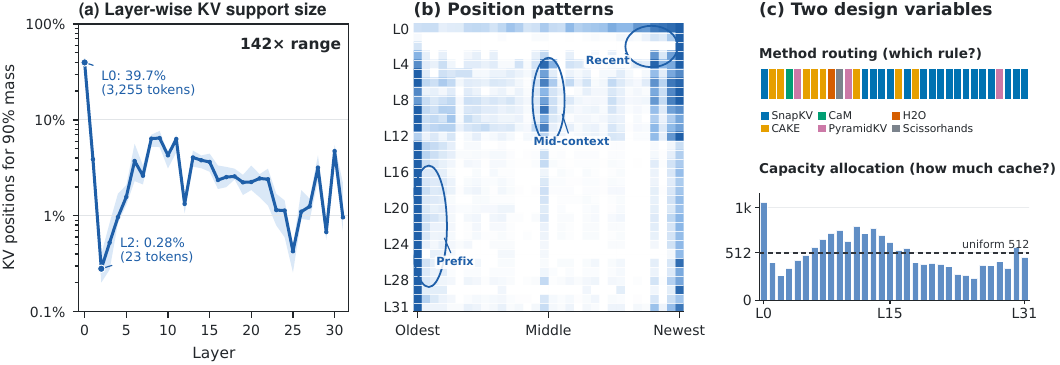}
\caption{Layer-wise attention heterogeneity in LLaMA-3.1-8B-Instruct.
Across 50 LongBench inputs with 8K-token prompts, we identify at the first decoding step the smallest KV-position set covering 90\% of head-averaged attention.
(a) Median set size varies by $142\times$ across layers; shading shows stratified 95\% bootstrap intervals.
(b) Row-normalized 32-bin support density for a representative input; outlines mark recent-token, mid-context, and prefix-focused patterns.
(c) An evaluated RULER plan separates routing from allocation: upper-strip colors encode methods, lower bars encode entropy-derived capacities, and the dashed line marks uniform 512-token capacity.}
\label{fig:motivation_layer_method_budget}
\end{figure}

Recent work relaxes uniform cache management through layer- or head-wise capacity adaptation, head specialization, and layer-wise routing of compression parameters~\citep{fastgen,pyramidkv,adakv,dynamickv,lava,razorattention,duoattention,headkv,moend}.
Yet a complementary question remains largely unexplored: whether different standalone retention-rule families should be assigned to different layers.
Transformer layers differ markedly in attention patterns, representations, contextual roles, and pruning sensitivity~\citep{tenney2019bert,jawahar2019bert,clark2019bert,voita2019analyzing,michel2019sixteen}.
Figure~\ref{fig:motivation_layer_method_budget} shows sharp layer-wise variation in attention-support size and location, suggesting potential value in varying cache capacities and retention biases across layers.
However, this variation does not establish layer-specific method preferences, since a sufficiently adaptive shared rule may already capture it.
Model-wide evaluations therefore leave a question unanswered: \emph{Does every layer need to use the same KV eviction method?}

To answer this question, we study layer-wise KV cache eviction design along two axes: \emph{method routing}, which assigns a retention-rule family to each layer within the active phase or phases, and \emph{capacity allocation}, which distributes a fixed aggregate nominal retained-token capacity across layers within each active phase.
Our factorial comparisons separate routing, allocation, and their interaction under matched prefill/decode scope and nominal retained-token capacity.
Within routing, a count-matched control isolates placement by fixing method counts and varying only layer locations.

Within the layer-wise routing space, which grows exponentially with model depth, our objective is to determine whether retention-rule identity constitutes an exploitable layer-wise design dimension, rather than to identify a globally optimal route.
We use \textsc{PolyKV} to represent each candidate as a layer- and phase-specific plan over compatible eviction methods and cache capacities.
To construct tractable candidates, we profile existing methods on a small held-out set, using FullKV-preservation diagnostics for routing and layer-level proxies for allocation.
These signals separately instantiate method-routing and capacity-allocation candidates.
Once constructed, each plan remains fixed at inference and requires no retraining or online search.
This process yields reproducible candidates for the controlled comparisons that follow.

Our results provide complementary positive evidence for layer-wise method routing.
On LongBench, matched within-model routing contrasts are $+0.26$ points on LLaMA-3.1-8B-Instruct and $+0.27$ points in a retrospective exact-capacity Qwen3-8B check.
Non-uniform allocation yields smaller and mixed changes, identifying routing as the more strongly supported factor.
The count-matched placement control further tests whether layer locations matter after method counts are fixed.
The guided route scores 0.31 points above the mean of 99 random count-matched placements and ranks second among the 100 evaluated assignments.
One route held fixed from 128 to 1024 average tokens per layer improves over the budget-wise best standalone eviction baseline at all five operating points by 0.46 to 2.22 points.
The largest gains occur under stronger compression.
A second workload-specific construction on RULER improves over matched SnapKV by 15.98 points and remains ahead of that anchor at all four context lengths.
The gains vary sharply across task families and include a regression on word extraction.
The overall score remains slightly below standalone TOVA.
Taken together, these results show that layer-wise routing of independently developed standalone retention rules can add value beyond selecting a single strong homogeneous method.
Thus, even when the constituent rules are competitive on their own, method routing creates an independent design axis, with particularly strong evidence that routing quality depends on layer placement.

Our contributions are threefold.
\begin{itemize}
    
    \item We formulate two axes of layer-wise KV eviction: method routing and capacity allocation.

    \item We develop \textsc{PolyKV} for profile-guided construction and controlled evaluation of static layer-wise plans under matched nominal retained-token capacities.

    \item We show that method routing contains exploitable placement-sensitive structure, while the tested non-uniform allocations yield weaker and inconsistent effects.
\end{itemize}

\section{Related Work}
\label{sec:related_work}

\subsection{Retention Rules as Model-Wide Alternatives}

KV cache eviction reduces long-context inference memory by retaining or consolidating only a subset of historical key and value states.
Existing methods embody different assumptions about which states will remain useful.
H2O preserves heavy hitters using accumulated attention, SnapKV scores prompt positions through a recent observation window, Scissorhands tracks persistent importance across decoding steps, and TOVA uses attention from the current query~\citep{h2o,snapkv,scissorhands,tova}.
StreamingLLM instead follows a positional rule that retains attention sinks and recent tokens~\citep{streamingllm}.
CaM consolidates cached states through merging, while D2O combines cumulative-attention eviction with recall and merge operations~\citep{cam,d2o}.
These methods provide retention-rule families with meaningfully different biases toward historical importance, recency, position, and consolidation.

Applying one rule family throughout a model does not imply that every layer retains the same positions.
Layer-local attention signals can cause a shared rule to select different states at different layers.
The relevant convention is that the identity of the rule used to make those selections is commonly fixed across layers.
Representative formulations and evaluations therefore treat retention-rule families as standalone model-wide alternatives.

\subsection{Component-Specific KV Cache Management}

Adaptive allocation relaxes a different uniformity assumption by changing how much cache each component receives.
PyramidKV uses a pyramidal layer-budget schedule, while CAKE derives layer preferences from spatial and temporal attention dynamics~\citep{pyramidkv,cake}.
D2O also incorporates layer-dependent capacity within its own eviction framework~\citep{d2o}.
Ada-KV reallocates capacity across heads and can augment different base eviction policies, while HeadKV assigns head-specific budgets using retrieval and reasoning importance~\citep{adakv,headkv}.
DynamicKV updates layer-wise retained-token counts, and LAVa derives a unified information-loss-based eviction metric together with dynamic head and layer budgets~\citep{dynamickv,lava}.
These methods show that layers and heads need not receive uniform capacity.
Even when scoring and allocation are designed jointly, however, the retention decisions generally remain governed by a shared method-specific rule family.

A related line of work makes cache behavior heterogeneous across attention heads.
FastGen profiles heads and selects head-level cache behaviors according to their attention patterns~\citep{fastgen}.
RazorAttention and DuoAttention distinguish retrieval heads from local or streaming heads, preserving richer history for retrieval heads while compressing the remaining heads more aggressively~\citep{razorattention,duoattention}.
These approaches establish the value of component-specific policy behavior.
Their choice spaces are defined by framework-specific head categories rather than by layer-wise routing among independently proposed eviction methods.

MoE-nD brings heterogeneous assignment to the layer level through multi-axis compression routing~\citep{moend}.
It assigns each layer a tuple consisting of an eviction ratio, key precision, and value precision under a global budget.
The routed object is therefore a compression configuration rather than the identity of an independently developed retention-rule family.
This separates compression-parameter variation from variation in the retention algorithm.

Together, these studies establish the value of component-specific KV cache management through heterogeneous capacity allocation, head-level cache-behavior specialization, and layer-wise routing of compression parameters.
This work studies a complementary axis, namely whether independently developed standalone eviction-rule families can be routed across Transformer layers.
We use a common experimental interface to represent method identity and cache capacity as separate layer-wise factors, construct static layer- and phase-specific candidate plans, and evaluate method routing and capacity allocation under matched nominal retained-token capacity.
Prefill and decoding define the deployment scope of each comparison.

\section{Layer-Wise Method Routing and Capacity Allocation}
\label{sec:methodology}

\begin{figure}[t]
    \centering
    \includegraphics[width=\linewidth]{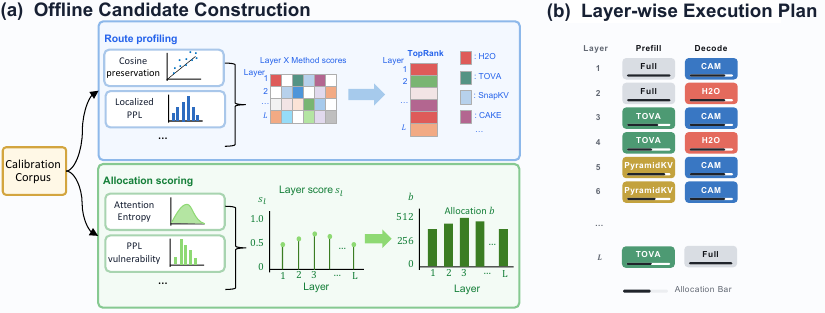}
    \caption{Overview of the controlled-plan construction.
    (a) Offline profiles from a calibration corpus produce layer-wise method rankings and allocation scores.
    (b) The resulting static execution plan assigns an operator and nominal capacity to each layer and inference phase; \textsc{Full} denotes no additional eviction in that phase.}
    \label{fig:polykv_methodology_overview}
\end{figure}

We ask whether eviction-method identity is a meaningful layer-wise design variable for KV cache compression.
Heterogeneous plans may differ in routing, allocation, deployment phase, or total budget; routes may also differ in method counts or layer placement.
We therefore factorize the space and compare matched configurations.
Offline profiles construct reproducible candidates in the exponentially large routing space, as summarized in Figure~\ref{fig:polykv_methodology_overview}.
The plans are frozen before inference, and \textsc{PolyKV} executes them through a common interface and runtime; it introduces neither a new retention rule nor an online route selector.

\subsection{Factorized Design Space and Controlled Contrasts}
\label{sec:method_formulation}

Consider an autoregressive Transformer with $L$ layers and prefill and decoding phases $\Phi=\{\mathrm{pre},\mathrm{dec}\}$.
Let $\mathcal{R}_{\phi}$ be the operators compatible with phase $\phi$.
Some compress the prompt cache after prefill; others update or sparsely access it during decoding.
A layer-wise execution plan is
\begin{equation}
    \Pi
    =
    \left\{
    \left(
    \{r_{l,\phi}\}_{\phi\in\Phi},
    \{b_{l,\phi}\}_{\phi\in\Phi}
    \right)
    \right\}_{l=1}^{L},
    \label{eq:polykv_plan}
\end{equation}
where $r_{l,\phi}\in\mathcal{R}_{\phi}\cup\{\textsc{Full}\}$ is the operator at layer $l$ in phase $\phi$.
For a capacity-controlled operator, $b_{l,\phi}$ is its nominal retained-token capacity; otherwise, it is undefined.
An inactive phase uses \textsc{Full} and performs no additional phase-specific eviction.
Figure~\ref{fig:polykv_methodology_overview}(b) shows this plan.

Deployment scope matters because prefill determines which prompt KV states reach decoding; decoding then accesses and updates them during generation.
We denote a route's active phases by $\delta\in\{\mathrm{prefill},\mathrm{decode},\mathrm{both}\}$.
Although Equation~\ref{eq:polykv_plan} permits different operators across phases, we evaluate prefill-only, decoding-only, and restricted both-phase routes.
In a both-phase route, each layer uses the same routed operator in both phases, chosen from operators compatible with both.

For comparisons with nominal capacities at all active assignments, matched configurations fix the total $B_{\phi}=\sum_{l=1}^{L}b_{l,\phi}$ and average $\bar b_{\phi}=B_{\phi}/L$ within each active phase.
This matches configured retained KV positions, not measured memory, auxiliary state, latency, or runtime.

Under a fixed model, workload, deployment scope, and total nominal capacity, we compare four families.
Homogeneous--uniform uses one operator and capacity across layers; allocation-only redistributes capacity without changing operator assignments; route-only varies operators under uniform capacity; and joint adds non-uniform allocation to the same heterogeneous route.
Thus, routing is compared at fixed allocation and allocation at fixed operator assignments; the factorial comparison measures their interaction.
The homogeneous--route contrast changes both method mixture and layer assignment, so it cannot alone identify a placement effect.

To isolate placement, define a completed route and its operator count as
\[
    \rho_{\delta}=\{\rho_{l,\delta}\}_{l=1}^{L},
    \qquad
    N_r(\rho_{\delta})=|\{l:\rho_{l,\delta}=r\}|.
\]
A count-matched placement preserves every $N_r(\rho_{\delta})$ and changes only which layers receive each operator.
We compare the guided assignment with 99 unique random assignments from this space, excluding the guided assignment, while holding the model, workload, scope, allocation, and nominal budget fixed.
The guided assignment is fixed before sampling and scoring.
This control tests for placement information beyond the method multiset, not global optimality.

For the budget sweep, we hold a completed route's operator identities and layer assignments fixed while varying $\bar b_{\phi}$, without reselecting the route.
This traces one construction's quality--capacity behavior rather than a budget-wise oracle assembled from different routes.

\subsection{Profile-Based Route Construction}
\label{sec:method_calibration}

For scope $\delta$, let $\mathcal{R}_{\delta}$ contain the operators compatible with every active phase; this space contains $|\mathcal{R}_{\delta}|^L$ assignments, making exhaustive evaluation infeasible.
We instead use a heterogeneous calibration corpus $\mathcal{D}_{\mathrm{cal}}$ comprising 20 sequences, each at most 2{,}048 tokens, from WikiText, C4, PG-19, and CodeParrot.
FullKV supplies the reference behavior, and candidate operators are profiled at the common calibration capacity $b_{\mathrm{cal}}=256$.
The resulting statistic $g_{l,\delta,r}$ ranks compatible operators at each layer.
The deterministic profile-to-route mapping does not use scores from the placement controls, and the designated route is frozen before those controls are run.

The upper branch of Figure~\ref{fig:polykv_methodology_overview}(a) shows two route profiles: representation similarity and perplexity degradation.
For representation similarity, operator $r$ is applied to every layer within scope $\delta$.
During free-running continuation generation, we pool each layer's attention-module outputs over the generated trajectory and calibration corpus.
The cosine similarity between the resulting FullKV and candidate summaries is
\[
    C_{l,\delta,r}
    =
    \cos\!\left(\bar h^{\mathrm{full}}_l,\bar h^{(\delta,r)}_l\right),
\]
where larger values indicate closer pooled representations.
Because $r$ acts scope-wide and deeper layers may inherit upstream compression effects, this layer-indexed diagnostic reflects the intervention, not a layer-local causal preference.

The localized PPL profile covers the fixed set $\mathcal{R}_{\mathrm{profile}}$: \textsc{Cake}, \textsc{Cam}, \textsc{D2O}, \textsc{H2O}, \textsc{PyramidKV}, \textsc{Scissorhands}, \textsc{SnapKV}, and \textsc{Tova}.
For each operator $r$, only layer $l$ uses it in both prefill and decoding; all other layers retain FullKV.
On teacher-forced continuation suffixes, we define
\[
    \Delta\mathrm{PPL}_{l,r}
    =
    \mathrm{PPL}_{l,r}-\mathrm{PPL}_{\mathrm{full}},
\]
where smaller values indicate closer FullKV predictions.
For a requested scope, we restrict this phase-shared profile to operators compatible with that scope; it does not estimate separate prefill and decoding preferences.
Profiles are not combined; each yields one route.

For profile $g$, let $\mathcal{R}_{\delta}^{g}\subseteq\mathcal{R}_{\delta}$ denote the compatible operators covered by $g$.
The constructor selects
\begin{equation}
    \rho_{l,\delta}
    =
    \operatorname{TopRank}_{g}
    \left(
    \{(r,g_{l,\delta,r})\mid r\in\mathcal{R}_{\delta}^{g}\}
    \right),
    \label{eq:policy_routing}
\end{equation}
where higher is better for cosine preservation and lower for PPL degradation.
The selected operator is installed in every active phase, while inactive phases use \textsc{Full}, yielding a prefill-only, decoding-only, or tied both-phase candidate.
This independent layer-wise rule ignores cross-assignment interactions.
We use each profile as a separate heuristic to construct a reproducible route, without searching for which profile best supports route selection.

\subsection{Capacity Allocation as a Separate Factor}
\label{sec:method_construction}

After fixing the route, the route-only plan assigns $b_{l,\phi}=\bar b_{\phi}$ to every capacity-controlled layer in each active phase.
Allocation refinement keeps all operator assignments fixed and redistributes the same total using layer score $s_l$.

The lower branch of Figure~\ref{fig:polykv_methodology_overview}(a) shows two allocation scores: attention entropy and compression-induced PPL degradation.
Let $A^{(l)}_{h,q,k}(x)$ denote the FullKV attention probability from query $q$ to key $k$ in head $h$ at layer $l$ on calibration sequence $x$, and let $\mathcal{H}$ denote Shannon entropy.
We define
\[
    H_l=\mathbb{E}_{x,h,q}\!\left[\mathcal{H}\!\left(A^{(l)}_{h,q,:}(x)\right)\right],
    \qquad
    S^{\mathrm{ppl}}_l=\max_{r\in\mathcal{R}_{\mathrm{profile}}}\Delta\mathrm{PPL}_{l,r}.
\]
Higher entropy indicates diffuse attention, so the heuristic assigns such layers more capacity; downstream evaluation tests whether this improves quality.
Larger $S^{\mathrm{ppl}}_l$ values indicate layers more affected by the profiled interventions, so the allocator gives them more capacity.
Because every operator is profiled at one capacity, this score measures compression vulnerability rather than marginal capacity sensitivity.
Both scores are model-specific, phase-shared, and fixed before evaluation.

Given a score $s_l$, a fixed transformation $T$ produces non-negative weights.
Entropy allocation uses the identity, whereas PPL-derived allocation caps the score at the across-layer 95th percentile:
\[
    T(H_l)=H_l,
    \qquad
    T(S^{\mathrm{ppl}}_l)=\min\!\left\{S^{\mathrm{ppl}}_l,Q_{0.95}(S^{\mathrm{ppl}})\right\}.
\]
The cap prevents one outlier from dominating the total.
If the transformed minimum is non-positive, all values are shifted by $-\min_l T(s_l)+10^{-6}$ before normalization.
The provisional allocation is
\begin{equation}
    w_l
    =
    \frac{T(s_l)}
    {\sum_{j=1}^{L}T(s_j)},
    \qquad
    \tilde b_{l,\phi}
    =
    B_{\phi}w_l.
    \label{eq:budget_allocation}
\end{equation}
The allocator iteratively fixes layers below $b_{\min}$ at that minimum and redistributes the remainder among other layers in proportion to their weights.
We use $b_{\min}=32$ for entropy allocation and $b_{\min}=48$ for PPL-derived allocation; the latter keeps SnapKV's capacity above its 32-token observation window.
After nearest-integer rounding, residual units are assigned by fractional remainder, so $b_{l,\phi}\ge b_{\min}$ and $\sum_{l=1}^{L}b_{l,\phi}=B_{\phi}$.
Applied to a homogeneous or heterogeneous route, the procedure yields an allocation-only or joint configuration without changing operator assignments.
For the non-uniform-allocation budget sweep, we scale the completed 512-token vector to each target total and apply largest-remainder rounding; neither the score profile nor the route is recomputed.

\subsection{Executing Controlled Plans with \textsc{PolyKV}}
\label{sec:method_interface}

To compare methods developed as standalone alternatives, \textsc{PolyKV} preserves their scoring and retention logic through static configuration, a common operator interface, and phase- and layer-specific dispatch.
Each adapter declares compatible phases and implements the handlers required by the original method for attention-kernel dispatch, prompt-cache compression, and incremental cache updates.
Capacity-controlled adapters also expose $b_{l,\phi}$.
These declarations define $\mathcal{R}_{\phi}$ and the intersection for tied both-phase candidates.
Outside declared handlers, execution follows the standard attention path, and inactive phases use \textsc{Full}.

Each completed plan is serialized and loaded once.
The runtime tracks the prefill-to-decoding transition and dispatches each layer according to Equation~\ref{eq:polykv_plan}.
Calibration is offline, and routing and allocation remain static during evaluation.
Execution requires no model-weight changes, online profiling, route search, or policy adaptation.

\begin{figure}[t]
\centering
\includegraphics[width=\linewidth]{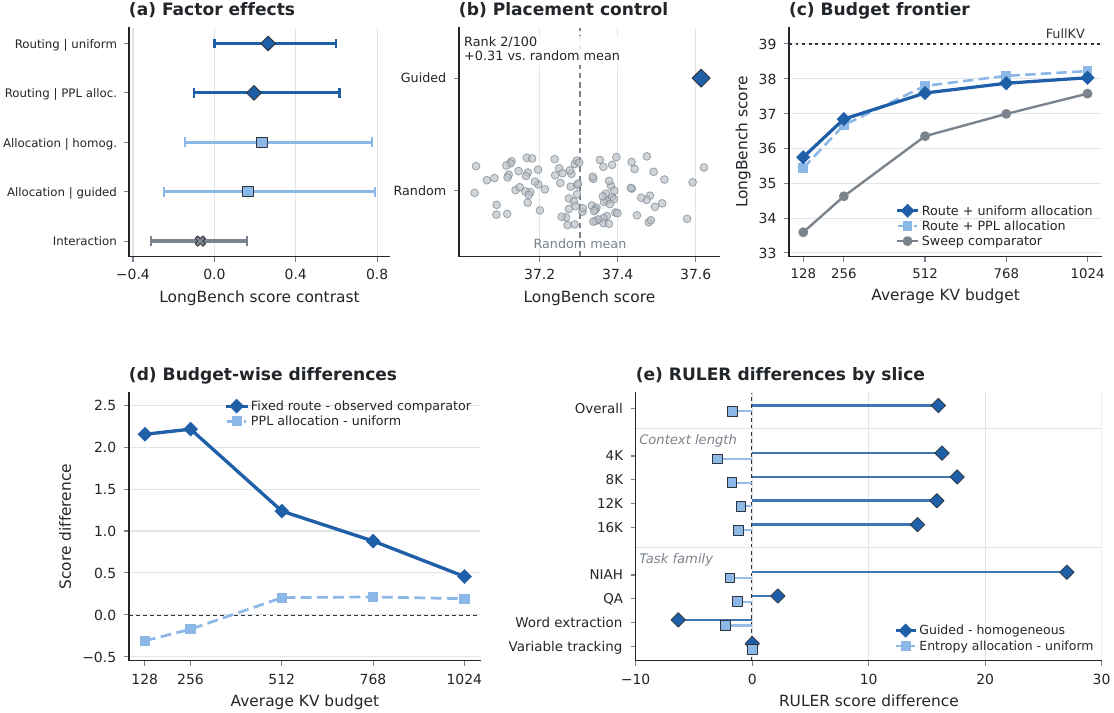}
\caption{Five views of profile-guided method routing and capacity allocation.
(a) Task-paired routing effects (guided cosine route minus homogeneous TOVA) under uniform and PPL-derived allocation, allocation effects (PPL-derived minus uniform) under homogeneous and guided routing, and their interaction on LongBench; bars are 95\% bootstrap intervals.
(b) LongBench scores of the frozen guided placement and 99 unique random placements with the same counts of three D2O and 29 TOVA layers.
(c) LongBench scores across five average KV budgets for one fixed route with uniform or PPL-derived allocation, the strongest observed standalone result among nine baselines at each budget, and FullKV.
(d) The uniform-allocation fixed route minus the budget-wise standalone reference, and PPL-derived minus uniform allocation within the fixed route; the reference is TOVA at 128 tokens and CAKE at 256--1024 tokens.
(e) RULER differences for the guided uniform policy minus homogeneous SnapKV under uniform allocation, and for entropy-derived minus uniform allocation within the guided route, reported overall and by context length and task family.
Only (a) reports bootstrap uncertainty; panels (b)--(e) are descriptive.}
\label{fig:evaluation_evidence}
\end{figure}

\section{Evaluation}

The evaluation addresses three questions.
First, under a matched nominal KV budget, can evidence for method routing be separated from evidence for non-uniform capacity allocation?
Second, after fixing the method mixture, does the offline profile provide an informative ranking of sampled layer placements?
Third, does favorable construction-level behavior recur across capacity constraints and on a second model, and what workload-specific boundaries appear?
Our aim is not to identify a global optimum or an automatic route selector.
We therefore distinguish primary controlled evidence on LongBench, supporting descriptive consistency across budgets and a second model, and a workload-specific RULER boundary.
This organization keeps the conclusion narrow: the tested routing and placement space contains exploitable profile-guided structure, whereas the tested allocation refinements and reliable selection on unseen workloads remain unresolved.

\subsection{Experimental Setup and Evidence Roles}

The primary controlled study uses LLaMA-3.1-8B-Instruct~\citep{llama3} on LongBench~\citep{longbench} and reports the macro average over 21 tasks.
A retrospective Qwen3-8B~\citep{qwen3} check uses the same 21-task LongBench macro protocol to assess a uniquely closest exact-capacity pair. The check is a targeted comparison with a narrower scope than a second full evaluation.
The RULER~\citep{ruler} case uses LLaMA-3.1-8B-Instruct and reports the official average over 13 tasks, four context lengths, and 52 task--length cells.

The framework integrates nine representative KV eviction methods.
They are H2O, TOVA, CaM, PyramidKV, SnapKV, D2O, CAKE, Scissorhands, and StreamingLLM.
FullKV provides the uncompressed reference.
Every compressed comparison matches the nominal average retained-token budget within its protocol.
This criterion controls configured retained KV positions and does not imply matched memory, auxiliary state, latency, or throughput.
Appendix~\ref{app:system_metrics} reports measured steady KV-related memory and prefill/decode throughput for one fixed H2O/TOVA composition.

The LongBench factorial and placement studies use a cosine decode route frozen before the controlled runs, while the original-cohort sweep evaluates it across budgets.
These studies provide the primary controlled evidence and the within-route capacity check, respectively.
The Qwen comparison provides a retrospective second-model check, whereas RULER uses a separately constructed workload-specific prefill policy to expose a workload boundary.
Appendix Table~\ref{tab:evaluation_protocol_registry} summarizes settings and evidence roles.
Selection-free route choice on unseen workloads remains unresolved.

\subsection{Controlled LongBench Evidence for Routing and Placement}
\label{sec:longbench_matched_controls}

We first ask whether a frozen profile-guided route improves over a homogeneous route under matched allocation rules, and whether non-uniform allocation provides an independent benefit.
All four cells use decode-only compression and an average budget of 512 tokens per active layer.
The routing factor compares homogeneous TOVA with the guided cosine route.
The allocation factor compares uniform capacity with a PPL-derived allocation under the same total nominal capacity.
Figure~\ref{fig:evaluation_evidence}a reports the task-paired contrasts.
Appendix Table~\ref{tab:appendix_longbench_factorial} gives the complete cells and intervals.

Only the routing contrast under uniform allocation has an interval that excludes zero: the guided route gains 0.2638 points with a task-bootstrap interval of $[+0.0005,+0.5976]$, which only narrowly excludes zero.
For this contrast, guided-route scores are higher on 11 tasks, unchanged on 2, and lower on 8.
Despite this heterogeneity, all 21 leave-one-task-out macro contrasts remain positive, ranging from $+0.1285$ to $+0.3150$.
Together, task-level variation and leave-one-out stability provide modest macro-average evidence that the routing profile is informative relative to the homogeneous anchor, not that improvement is task-universal.
Other factorial intervals include zero, leaving the tested allocation refinement unresolved. Appendix Table~\ref{tab:appendix_longbench_task_deltas} lists all paired differences.

The routing contrast changes method mixture and placement together.
We therefore fix method counts at three D2O and 29 TOVA layers, changing only the D2O locations.
The model, workload, decode-only scope, uniform allocation, and 512-token average budget remain fixed.
Figure~\ref{fig:evaluation_evidence}b compares the guided assignment with 99 unique random placements from the count-matched space.

Because the guided assignment was fixed before sampling, its upper-tail position yields a corrected one-sided Monte Carlo value of $p_{\mathrm{MC}}=0.020$.
Appendix Table~\ref{tab:appendix_route_placements} summarizes the sampled placement distribution.
The finite sample does not identify the best of all possible placements; instead, the guided assignment's position relative to sampled count-matched alternatives provides evidence against a wholly uninformative placement signal.

\paragraph{Profile validation.}
Across the same 99 random placements, the frozen offline profile score is positively associated with downstream LongBench performance (tie-aware Spearman $\rho=0.459$; corrected one-sided permutation $p=0.0001$).
Excluding the guided assignment, this association shows that the profile ranks performance within the sampled count-matched placement space, beyond one strong route.
Appendix Figure~\ref{fig:appendix_profile_placement_validation} shows this relationship.

Together, the matched routing gain, upper-tail fixed placement, and profile--performance association across sampled count-matched placements provide the primary evidence for exploitable routing and placement structure.
They do not identify a global optimum or imply task-universal improvement.

\subsection{Supporting Consistency Across Capacity Constraints and a Second Model}
\label{sec:fixed_route_frontier}

We next test capacity consistency without route reselection by holding the cosine decode route and its layer assignments fixed from 128 to 1024 average tokens per layer.
At every budget, the original sweep cohort evaluates the same nine standalone eviction baselines, so their per-budget maximum provides a clearly delimited strongest observed reference rather than an estimate of the global best.
Figures~\ref{fig:evaluation_evidence}c and~\ref{fig:evaluation_evidence}d report scores and differences.

Under uniform allocation, the fixed route exceeds the strongest observed standalone score at all five matched budgets, with margins ranging from $+0.46$ to $+2.22$ points.
At 128, 256, and 512 tokens, it also matches or exceeds the strongest standalone reference at twice the nominal retained capacity.
This comparison does not measure memory, latency, or throughput.
Because its layer assignments are not reselected, this behavior across the five tested constraints supports construction-level consistency rather than success at one operating point.
By contrast, the changing direction of the PPL-minus-uniform effect across budgets does not provide the tested allocation heuristic with the same consistent support as routing.
Appendix Table~\ref{tab:appendix_llama_longbench_budget_sweep_matrix} gives the full result matrix.

\paragraph{Second-model descriptive consistency.}
The retrospective audit rule in Appendix~\ref{app:qwen_cross_model_check} identifies a Qwen3-8B pair with FullKV prefill, decode-only compression, and identical entropy-derived layer capacities with mean 512.
The homogeneous H2O anchor scores 46.9619, while the routed variant, which changes two decode assignments from H2O to TOVA, scores 47.2333, a $+0.2714$ difference.
The routed plan improves 13 tasks, ties on 3, and declines on 5. The median difference is $+0.1400$, and the descriptive paired-bootstrap interval is $[+0.0148,+0.5553]$.
Because the audit rule was formalized retrospectively, this similar direction and magnitude provide descriptive second-model consistency for a within-model LongBench contrast, not confirmatory replication or transfer of a common route.
The same exact pair changes the RULER average by only $-0.0712$; this observed near-zero difference does not support cross-workload transfer.

\subsection{A Workload-Specific Construction and Its Boundary on RULER}
\label{sec:ruler_second_case}

RULER tests whether a separately profiled, workload-specific construction can also reach a high-performing range under a different protocol; it does not test transfer of the LongBench route.
It uses prefill-only compression at a 512-token average budget.
Because SnapKV occupies 18 of the 32 layers in the frozen route, the factorial comparison uses homogeneous SnapKV as the dominant-method ablation anchor. Standalone TOVA is reported separately as the strongest homogeneous performance reference.
Figure~\ref{fig:evaluation_evidence}e reports overall, length, and task-family differences for the frozen protocol-matched comparisons.
Appendix Table~\ref{tab:appendix_ruler_factorial} reports all four cells and adds one newly completed homogeneous entropy cell; effects involving that cell are interpreted descriptively.

The large matched-anchor gain does not make the guided uniform construction the best observed policy: it is about 16 points above homogeneous SnapKV but 0.10 points below standalone TOVA overall.
NIAH accounts for 32 of the 52 task--length cells and drives most of the aggregate improvement.
Variable tracking remains at the floor under both the guided-uniform and homogeneous-SnapKV policies, so its zero contrast does not indicate robust parity.
Together with the cross-family directions in Figure~\ref{fig:evaluation_evidence}e, this concentration makes the matched-anchor advantage task-family-specific rather than task-universal.
The non-positive entropy-minus-uniform differences in the non-floor slices provide no supporting evidence for the tested allocation signal in this protocol.

Taken together, the primary LongBench controls support exploitable structure in method routing and placement, including an association between the frozen offline profile and performance within the sampled count-matched placement space.
The fixed-route budget sweep and retrospective Qwen check add supporting construction-level consistency, but neither establishes automatic route selection or cross-workload transfer.
The near-zero Qwen RULER difference and the workload-specific RULER case delimit the claim, while the tested allocation refinements remain unresolved.
These experiments neither identify a global optimum nor establish universal superiority.

\section{Conclusion}

This work asks whether KV eviction-method identity should remain a model-wide choice.
Our results indicate that it need not: in the tested settings, method routing is a meaningful layer-wise design dimension, with the clearest evidence coming from placement-sensitive structure.

The primary matched LongBench comparison provides modest macro-average support for routing over a homogeneous anchor, while the tested non-uniform allocation refinements remain unresolved.
With method counts fixed, the guided assignment ranks second among 100 count-matched placements, and the frozen offline profile is positively associated with LongBench performance across that sampled space.
A fixed route supports construction-level consistency by exceeding the strongest observed score among the same nine standalone baselines at all five matched nominal budgets.
A retrospective Qwen comparison is a separate descriptive second-model check with a similar direction on LongBench.
The near-zero Qwen RULER difference and the task-family concentration of a separately constructed RULER route limit cross-workload and task-universal claims.

These results do not identify a globally optimal route or establish reliable selection on unseen workloads.
The separate system characterization in Appendix~\ref{app:system_metrics} covers steady KV-related memory, prefill throughput, and decode throughput for one fixed composition and does not establish end-to-end gains across routes or workloads.
The evidence shows KV eviction methods should be studied as both model-wide alternatives and components whose identity and placement can vary by layer.

\newpage

\subsection*{AI Use Statement}

Generative AI tools were used to assist with software-code implementation and
with editing the manuscript for clarity and readability. AI-assisted code was
reviewed and tested by the authors, and all AI-assisted manuscript changes were
reviewed by the authors. The authors take responsibility for the correctness of
the implementation and for the final content of this work, including any text,
code, or other artifacts produced with the aid of generative AI.

\subsection*{Reproducibility Statement}

Sections~\ref{sec:method_formulation}--\ref{sec:method_interface} define the
factorized plan, profiling signals, fixed-total allocator, and common execution
interface. Section~\ref{sec:longbench_matched_controls} and
Appendices~\ref{app:evaluation_protocol_registry}--\ref{app:complete_ruler_factorial}
document the evaluation settings, controlled comparisons, task-level results,
placement analysis, and the Qwen and RULER boundary checks.

\newpage

\bibliography{references}
\bibliographystyle{iclr2027_conference}

\newpage

\appendix
\raggedbottom

\section{Experimental Setup and Statistical Procedures}
\label{app:evaluation_protocol_registry}

This appendix follows the evidence roles in the main paper.
We first summarize the evaluation settings and statistical procedures, then report the primary LongBench controls, supporting consistency across cache budgets and a second model, the RULER boundary analysis, and a system-level characterization of one fixed composition.
Every inactive inference phase uses FullKV.

\begin{table}[H]
\caption{Summary of the evaluation settings and evidence roles. All compressed comparisons match total nominal retained-token capacity within their stated protocol.}
\label{tab:evaluation_protocol_registry}
\centering
\CompactTable
\begin{tabularx}{\linewidth}{@{}
>{\raggedright\arraybackslash}p{0.14\linewidth}
>{\raggedright\arraybackslash}p{0.23\linewidth}
>{\raggedright\arraybackslash}p{0.20\linewidth}
>{\raggedright\arraybackslash}p{0.17\linewidth}
Y@{}}
\toprule
Study & Evaluation setting & Route or anchor & Allocation and mean capacity & Evidence role \\
\midrule
Matched factorial & LLaMA-3.1-8B-Instruct / LongBench; decode & Guided cosine versus homogeneous TOVA & Uniform or PPL-derived; mean 512 & Separates routing, allocation, and their interaction \\
Count-matched placements & LLaMA-3.1-8B-Instruct / LongBench; decode & Three D2O and 29 TOVA assignments & Uniform; mean 512 & Isolates layer placement at fixed method counts \\
Retrospective cross-model check & Qwen3-8B / LongBench; decode & Two-layer H2O/TOVA route versus homogeneous H2O & Same entropy-derived vector; mean 512 & Tests within-model consistency under an exact capacity match \\
Fixed-route sweep & LLaMA-3.1-8B-Instruct / LongBench; decode & One frozen cosine route & Uniform or scaled PPL-derived; mean 128--1024 & Tests one construction across capacity constraints \\
Matched RULER case & LLaMA-3.1-8B-Instruct / RULER; prefill & Guided PPL route versus homogeneous SnapKV & Uniform or entropy-derived; mean 512 & Tests a second construction and localizes its task-family boundary \\
\bottomrule
\end{tabularx}
\end{table}

\FloatBarrier

\paragraph{Statistical procedures.}
All evaluated plans are constructed offline and remain fixed during inference; no route search or capacity adaptation uses downstream evaluation scores.
For the LongBench placement control, the guided assignment is fixed before evaluating 99 unique random count-matched assignments from the remaining placement space.
For profile validation, each placement receives the frozen score
$S(\rho)=\sum_{\ell\in D(\rho)}(C_{\ell,\mathrm{D2O}}-C_{\ell,\mathrm{TOVA}})$.
Here, $D(\rho)$ denotes the layers assigned D2O by placement $\rho$, and $C_{\ell,r}$ is the frozen cosine-preservation score for method $r$ at layer $\ell$.
The primary statistic is a tie-aware Spearman rank correlation between $S(\rho)$ and the LongBench macro score over the 99 random placements.
Its corrected one-sided permutation value is computed by shuffling the 99 downstream scores 10{,}000 times.
The guided assignment is excluded from the primary calculation and included only in the secondary descriptive analysis.
For the LongBench factorial, confidence intervals use 10{,}000 paired bootstrap replicates over the 21 LongBench tasks.

\section{Complete Controlled LongBench Evidence}
\label{app:matched_longbench_controls}

The factorial below separates the guided routing policy from PPL-derived capacity redistribution.
The routing contrast changes both method counts and layer locations, whereas the subsequent placement control holds method counts fixed at three D2O and 29 TOVA layers and changes only their locations.

\subsection{Routing and Allocation}

Cell values are 21-task macro averages, and intervals use 10{,}000 paired bootstrap replicates over tasks.
Routing contrasts are guided minus homogeneous, while allocation contrasts are PPL-derived minus uniform.

\begin{table}[H]
\caption{Complete matched LongBench factorial for LLaMA-3.1-8B-Instruct at a 512-token average KV budget.}
\label{tab:appendix_longbench_factorial}
\centering
\StandardTable
\begin{tabular}{@{}lrr@{}}
\toprule
Routing policy & Uniform allocation & PPL allocation \\
\midrule
Homogeneous TOVA (decode-only) & 37.3500 & 37.5833 \\
Guided cosine route & 37.6138 & 37.7776 \\
\bottomrule
\end{tabular}

\vspace{0.45em}
\begin{tabular}{@{}lrr@{}}
\toprule
Contrast & Estimate & Task-bootstrap 95\% CI \\
\midrule
Routing under uniform & +0.2638 & $[+0.0005,+0.5976]$ \\
Routing under PPL & +0.1943 & $[-0.1010,+0.6143]$ \\
Allocation under homogeneous & +0.2333 & $[-0.1433,+0.7762]$ \\
Allocation under guided & +0.1638 & $[-0.2490,+0.7876]$ \\
Routing $\times$ allocation & -0.0695 & $[-0.3105,+0.1586]$ \\
\bottomrule
\end{tabular}
\end{table}

\paragraph{Task-level robustness.}
For the uniform-allocation routing contrast, the 21 paired task differences have mean $+0.2638$ and median $+0.0400$; guided routing improves 11 tasks, is unchanged on 2, and declines on 8.
All 21 leave-one-task-out macro contrasts remain positive, ranging from $+0.1285$ to $+0.3150$.
The largest individual contribution is \texttt{passage\_retrieval\_zh} at $+2.97$, which accounts for 53.6\% of the summed task-level contrast.
Omitting this task reduces the macro improvement to $+0.1285$ but does not reverse its direction.
Thus, the magnitude is concentrated, while no single task overturns the positive macro-average contrast.

\begin{table}[H]
\caption{Complete task-level differences for routing under uniform allocation in the matched LongBench study. $\Delta_{\mathrm{route}}$ is the guided cosine route minus homogeneous TOVA; positive values favor guided routing.}
\label{tab:appendix_longbench_task_deltas}
\centering
\CompactTable
\begin{tabular*}{\linewidth}{@{\extracolsep{\fill}}lrlr@{}}
\toprule
Task & $\Delta_{\mathrm{route}}$ & Task & $\Delta_{\mathrm{route}}$ \\
\midrule
\texttt{narrativeqa}             & $+0.72$ & \texttt{vcsum}                  & $+0.23$ \\
\texttt{qasper}                  & $+0.69$ & \texttt{trec}                   & $+0.00$ \\
\texttt{multifieldqa\_en}        & $+0.81$ & \texttt{triviaqa}               & $+0.44$ \\
\texttt{multifieldqa\_zh}        & $+0.73$ & \texttt{samsum}                 & $+0.04$ \\
\texttt{hotpotqa}                & $+0.32$ & \texttt{lsht}                   & $+0.00$ \\
\texttt{2wikimqa}                & $-0.02$ & \texttt{passage\_count}         & $+0.29$ \\
\texttt{musique}                 & $-0.40$ & \texttt{passage\_retrieval\_en} & $-0.76$ \\
\texttt{dureader}                & $+0.29$ & \texttt{passage\_retrieval\_zh} & $+2.97$ \\
\texttt{gov\_report}             & $-0.15$ & \texttt{lcc}                    & $-0.08$ \\
\texttt{qmsum}                   & $-0.23$ & \texttt{repobench-p}            & $-0.31$ \\
\texttt{multi\_news}             & $-0.04$ &                                 &         \\
\bottomrule
\end{tabular*}
\end{table}

\subsection{Count-Matched Placement and Profile Validation}

\begin{table}[H]
\caption{Summary of the count-matched placement control on LongBench. Every assignment contains three D2O layers and 29 TOVA layers under uniform allocation at a 512-token average budget. The control pool contains 99 unique random assignments and excludes the guided assignment.}
\label{tab:appendix_route_placements}
\centering
\CompactTable
\begin{tabular}{@{}lr@{}}
\toprule
Statistic & Value \\
\midrule
Random controls & 99 \\
Guided score & 37.6138 \\
Random mean (sample SD) & 37.3034 (0.1303) \\
Random range & [37.0324, 37.6210] \\
Guided minus random mean & $+0.3105$ \\
Random $\geq$ guided & 1 \\
Guided rank & 2/100 \\
\bottomrule
\end{tabular}
\end{table}

\begin{figure}[H]
\centering
\includegraphics[width=0.62\linewidth]{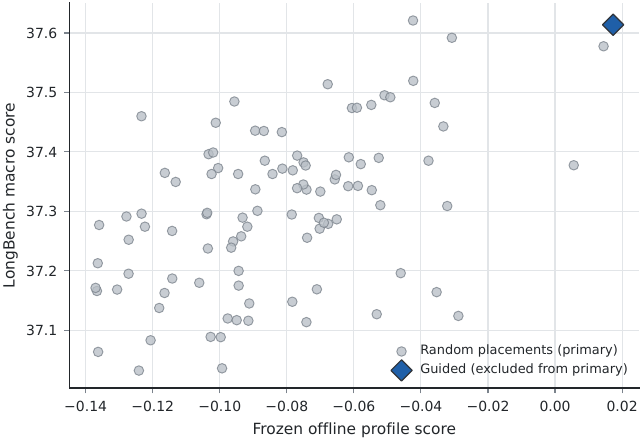}
\caption{Frozen-profile score versus downstream quality for count-matched placements. Across the 99 random placements, the primary tie-aware Spearman coefficient is $\rho=0.459$ with corrected one-sided permutation $p=0.0001$. The guided placement is shown separately and excluded from that calculation; including it gives the secondary coefficient $\rho=0.475$.}
\label{fig:appendix_profile_placement_validation}
\end{figure}
\FloatBarrier

\section{Supporting Consistency Across Budgets and Models}
\label{app:supporting_consistency}

\subsection{Fixed-Route Robustness Across Cache Budgets}
\label{app:longbench_budget_sweep}

The complete matrix reports 21-task macro averages from raw LongBench outputs; higher is better.
It holds the guided cosine route and its layer assignments fixed from 128 to 1024 average tokens per layer.
The uniform variant changes only the common capacity, while the PPL-derived variant scales the fixed 512-token profile to the target total.
FullKV is budget-free and shown once, while Quest is excluded because it is not an eviction method.
The same nine standalone eviction baselines appear at every budget under their original sweep protocols; in particular, TOVA is active in both prefill and decode, whereas the matched factorial in Table~\ref{tab:appendix_longbench_factorial} uses decode-only TOVA.
Their per-budget maximum defines the baseline envelope used in the main-text comparison.

\begin{table}[H]
\caption{Complete LLaMA-3.1-8B-Instruct LongBench budget sweep across five mean nominal capacities.}
\label{tab:appendix_llama_longbench_budget_sweep_matrix}
\centering
\CompactTable
\begin{tabular}{@{}lrrrrr@{}}
\toprule
Method / configuration & 128 & 256 & 512 & 768 & 1024 \\
\midrule
FullKV & \multicolumn{5}{c}{38.9857} \\
\midrule
\multicolumn{6}{@{}l}{\textit{Single baselines}} \\
CAKE & 32.1214 & 34.6248 & 36.3514 & 36.9890 & 37.5705 \\
CaM & 29.5862 & 31.4490 & 32.7248 & 33.9633 & 34.5600 \\
D2O & 11.9476 & 11.5695 & 11.6457 & 11.9238 & 11.8486 \\
H2O & 29.0733 & 31.3352 & 32.7910 & 33.9533 & 34.7167 \\
PyramidKV & 31.7910 & 34.1567 & 35.8090 & 36.6376 & 37.1414 \\
Scissorhands & 27.1529 & 29.2233 & 31.1881 & 32.0886 & 33.2905 \\
SnapKV & 32.2457 & 34.5805 & 35.9352 & 36.5167 & 36.9905 \\
StreamingLLM & 17.6043 & 16.8233 & 15.8124 & 15.5871 & 15.1319 \\
TOVA & 33.5924 & 34.4814 & 35.6962 & 36.4300 & 37.0571 \\
\midrule
\multicolumn{6}{@{}l}{\textit{Fixed layer-wise route variants}} \\
Route-only & 35.7476 & 36.8410 & 37.5890 & 37.8690 & 38.0267 \\
Route + PPL allocation & 35.4362 & 36.6681 & 37.7929 & 38.0814 & 38.2181 \\
\bottomrule
\end{tabular}
\end{table}

\FloatBarrier

\subsection{Retrospective Cross-Model Consistency on Qwen3-8B}
\label{app:qwen_cross_model_check}

We select the Qwen pair by identifiability rather than downstream rank.
Among 43 Qwen LongBench runs, the audit retains 22 Qwen-family-verified eviction recipes and eight exact-capacity non-routed--routed pairs.
The reported pair is the unique minimum-change comparison, differing in two decode assignments while keeping every layer capacity fixed.
Because the rule was formalized after the completed suite was inspected, this remains a retrospective consistency check.

On LongBench, the routed plan improves the macro average by 0.2714 points, with 13 wins, 3 ties, and 5 losses across the 21 tasks.
Its similar direction and magnitude to the matched LLaMA contrast provide descriptive construction-level consistency on a second model family, not confirmatory replication or transfer of a common policy.
The same pair is essentially neutral on RULER, restricting the evidence to within-model LongBench consistency and providing no support for cross-workload transfer.
In Table~\ref{tab:appendix_qwen_matched_summary}, arrows run from homogeneous to routed, and W/T/L counts routed wins, ties, and losses over 21 LongBench tasks or 52 RULER task--length cells.
The LongBench interval resamples 21 tasks, while the RULER interval resamples 13 task clusters across four context lengths; both use 10{,}000 bootstrap replicates.

\begin{table}[H]
\caption{Retrospective Qwen3-8B matched comparison and cross-workload boundary under identical layer-wise capacities.}
\label{tab:appendix_qwen_matched_summary}
\centering
\CompactTable
\begin{tabular*}{\linewidth}{@{\extracolsep{\fill}}lrrrr@{}}
\toprule
Workload & Matched scores & $\Delta$ & Descriptive 95\% CI & W/T/L \\
\midrule
LongBench & 46.9619 $\rightarrow$ 47.2333 & $+0.2714$ & $[+0.0148,+0.5553]$ & 13/3/5 \\
RULER & 24.3194 $\rightarrow$ 24.2483 & $-0.0712$ & $[-0.2315,+0.0812]$ & 14/21/17 \\
\bottomrule
\end{tabular*}
\end{table}

\FloatBarrier

\clearpage
\section{Matched RULER Case Study and Task-Family Boundary}
\label{app:complete_ruler_factorial}

The RULER study uses a separately constructed guided PPL-prefill route.
Table~\ref{tab:appendix_ruler_factorial} places its matched SnapKV anchor in the context of FullKV and standalone TOVA, then reports the complete routing-by-allocation cells, length slices, and task-family slices.
The factorial abbreviates homogeneous SnapKV as H, guided routing as G, uniform allocation as U, and entropy-derived allocation as E.
In the effect panel, $\Delta_R$ is guided minus homogeneous, $\Delta_E$ is entropy-derived minus uniform, and $\Delta_{R\times E}=\Delta_{R\mid E}-\Delta_{R\mid U}$.

The aggregate routing gain is concentrated in NIAH.
QA improves slightly, word extraction regresses, and variable tracking remains at the floor.
Entropy allocation lowers the guided score in every non-floor family and at every reported context length.
Three cells are frozen protocol-matched anchors; the homogeneous-entropy cell was completed separately, so effects involving that cell remain descriptive.

\begin{table}[H]
\caption{Reference context and complete RULER factorial scores and effects.}
\label{tab:appendix_ruler_factorial}
\centering
\CompactTable
\begin{tabular}{@{}lr@{}}
\toprule
Reference & Score \\
\midrule
FullKV & 83.01 \\
Homogeneous TOVA & 58.54 \\
\bottomrule
\end{tabular}

\vspace{0.55em}
\begin{tabular}{@{}lrrrr@{}}
\toprule
Slice & H--U & H--E & G--U & G--E \\
\midrule
All cells & 42.46 & 41.40 & 58.44 & 56.71 \\
\addlinespace
4K & 48.57 & 46.54 & 64.86 & 61.86 \\
8K & 40.95 & 40.10 & 58.54 & 56.79 \\
12K & 41.31 & 40.28 & 57.16 & 56.18 \\
16K & 39.03 & 38.69 & 53.21 & 52.02 \\
\addlinespace
NIAH & 44.76 & 43.25 & 71.77 & 69.84 \\
QA & 70.33 & 69.70 & 72.53 & 71.25 \\
Word extraction & 26.63 & 26.41 & 20.28 & 18.00 \\
Variable tracking & 0.00 & 0.00 & 0.00 & 0.00 \\
\bottomrule
\end{tabular}

\vspace{0.55em}
\begin{tabular}{@{}lrrrrr@{}}
\toprule
Slice & $\Delta_{R\mid U}$ & $\Delta_{R\mid E}$ & $\Delta_{E\mid H}$ & $\Delta_{E\mid G}$ & $\Delta_{R\times E}$ \\
\midrule
All cells & +15.98 & +15.31 & -1.06 & -1.73 & -0.67 \\
\addlinespace
4K & +16.30 & +15.32 & -2.03 & -3.01 & -0.98 \\
8K & +17.59 & +16.69 & -0.85 & -1.76 & -0.90 \\
12K & +15.85 & +15.90 & -1.03 & -0.98 & +0.05 \\
16K & +14.19 & +13.33 & -0.33 & -1.19 & -0.86 \\
\addlinespace
NIAH & +27.01 & +26.59 & -1.51 & -1.93 & -0.42 \\
QA & +2.20 & +1.55 & -0.63 & -1.28 & -0.65 \\
Word extraction & -6.36 & -8.41 & -0.22 & -2.28 & -2.05 \\
Variable tracking & 0.00 & 0.00 & 0.00 & 0.00 & 0.00 \\
\bottomrule
\end{tabular}
\end{table}

\clearpage

\section{System-Level Performance and Memory Characterization}
\label{app:system_metrics}

To complement the controlled quality evidence, we characterize one fixed H2O/TOVA composition at the implementation level.
The configuration uses LLaMA-3.1-8B-Instruct on a single NVIDIA H200 GPU with batch size 1, prompt lengths from 4K to 32K, and 128 generated tokens.
H2O is assigned to layers 0--15 and TOVA to layers 16--31, with every layer retaining 512 tokens.
H2O-512, TOVA-512, and \textsc{PolyKV} therefore retain 16{,}384 layer--token positions, corresponding to 64.00~MiB of exact KV state.
Steady memory includes the retained KV state and persistent eviction state and excludes model weights.
Table~\ref{tab:appendix_system_metrics} reports measured prefill throughput, decode throughput, and steady KV-related memory.

\begin{table}[H]
\caption{System-level measurements for the fixed H2O/TOVA composition on LLaMA-3.1-8B-Instruct.}
\label{tab:appendix_system_metrics}
\centering
\CompactTable
\begin{tabular*}{0.92\linewidth}{@{\extracolsep{\fill}}llrrr@{}}
\toprule
Context & Method & Prefill tok/s $\uparrow$ & Decode tok/s $\uparrow$ & Steady MiB $\downarrow$ \\
\midrule
4K  & FullKV             & 28{,}686.71 & 42.58 &   529.02 \\
4K  & H2O-512            & 10{,}113.81 & 25.59 &    65.14 \\
4K  & TOVA-512           & 17{,}130.04 & 29.58 &    65.14 \\
4K  & \textsc{PolyKV}    & 13{,}697.16 & 27.53 &    65.14 \\
\midrule
8K  & FullKV             & 27{,}289.50 & 42.26 & 1{,}041.14 \\
8K  & H2O-512            &  6{,}165.08 & 26.67 &    65.27 \\
8K  & TOVA-512           & 16{,}006.73 & 30.46 &    65.27 \\
8K  & \textsc{PolyKV}    &  8{,}962.82 & 27.91 &    65.27 \\
\midrule
16K & FullKV             & 22{,}886.05 & 40.31 & 2{,}065.39 \\
16K & H2O-512            &  3{,}193.75 & 26.69 &    65.52 \\
16K & TOVA-512           & 10{,}212.05 & 30.62 &    65.52 \\
16K & \textsc{PolyKV}    &  4{,}866.39 & 28.21 &    65.52 \\
\midrule
32K & FullKV             & 17{,}904.96 & 25.45 & 4{,}113.89 \\
32K & H2O-512            &  1{,}705.67 & 26.22 &    66.02 \\
32K & TOVA-512           &  6{,}054.22 & 30.08 &    66.02 \\
32K & \textsc{PolyKV}    &  2{,}662.11 & 26.05 &    66.02 \\
\bottomrule
\end{tabular*}
\end{table}

Across 4K--32K contexts, the fixed composition keeps measured steady memory at 65.14--66.02~MiB, whereas FullKV grows from 529.02 to 4{,}113.89~MiB.
The FullKV-to-compressed steady-memory ratio therefore grows from $8.1\times$ at 4K to $62.3\times$ at 32K.
Under the same 512-token-per-layer capacity, the same compose-consistent configuration achieves a LongBench macro average of 37.31 over 21 tasks, compared with 36.06 for both homogeneous H2O-512 and homogeneous TOVA-512.
These three quality scores share evaluation snapshot \texttt{12493d2} and contextualize only the fixed configuration in Table~\ref{tab:appendix_system_metrics}.
They are not pooled with later reruns.

The composition's runtime closely tracks the envelope of its constituent implementations rather than introducing a new dominant cost.
At every measured context length, its prefill throughput is higher than homogeneous H2O and lower than homogeneous TOVA.
Its decode throughput lies between the two through 16K and is within 0.7\% of H2O at 32K.
\textsc{PolyKV} prioritizes faithful composition through a common operator interface and layer- and phase-specific dispatch.
The current implementation does not use kernels specialized for a composed route, leaving fused dispatch and method-specific operations as direct optimization opportunities.
Thus, the current system-level benefit is higher LongBench quality at a bounded steady KV footprint, rather than universal latency acceleration.
\FloatBarrier

\end{document}